\documentclass[10pt,twocolumn,letterpaper]{article}

\usepackage{iccv}
\usepackage{times}
\usepackage{epsfig}
\usepackage{graphicx}
\usepackage{amsmath}
\usepackage{amssymb}
\usepackage{multirow}

\usepackage[pagebackref=true,breaklinks=true,letterpaper=true,colorlinks,bookmarks=false]{hyperref}

\iccvfinalcopy 

\def\iccvPaperID{****} 

\ificcvfinal\fi

\begin{document}
\renewcommand{\arraystretch}{1.2}

\title{Dropout Induced Noise for Co-Creative GAN Systems}

\author{Sabine Wieluch, Dr. Friedhelm Schwenker\\
Institute for Neural Information Processing\\
Ulm University\\
{\tt\small sabine.wieluch@uni-ulm.de, friedhelm.schwenker@uni-ulm.de}
}

\maketitle
\ificcvfinal\thispagestyle{empty}\fi

\begin{abstract}
   This paper demonstrates how Dropout can be used in Generative Adversarial Networks to generate multiple different outputs to one input. This method is thought as an alternative to latent space exploration, especially if constraints in the input should be preserved, like in A-to-B translation tasks. 
\end{abstract}

\section{Introduction}
 In current co-creative Generative Adversarial Network (GAN) systems, latent space exploration\cite{kingma2018glow} and manipulation\cite{zhu2016generative} is a common way to give the user a variety of possible generative outcomes. To give even more control to the user, neural net architectures like InfoGAN\cite{chen2016infogan} aim to learn disentangled latent space representations, so features of the generative model can be controlled separately.\\
Alas generating different outputs via latent space exploration is not a suitable solution for all generative settings. Many tasks require a generative system to start from a certain given input and not from a noise vector, for example Conditional GANs\cite{mirza2014conditional}. Typical examples for such tasks would be A-to-B translation\cite{zhu2017unpaired,isola2017image} like style transfer\cite{gatys2015neural}, image inpainting\cite{pathak2016context, yeh2017semantic} or image synthesis from text\cite{Zhang_2017_ICCV} or label masks\cite{reed2016learning}. In such cases, manipulations in the latent space could result in losing or altering the original constrains from the input vector.\\
One solution to this problem could be to feed an additional noise vector to the neural net, but Isola et al.\cite{isola2017image} and Mathieu et al.\cite{mathieu2015deep} describe that the generator only learns to ignore this noise.\\
Therefor we propose to use Dropout\cite{hinton2012improving} in the generation phase to create a variety of outputs.
	
\section{Dropout as induced Noise}
To receive multiple different results from one GAN input, we propose to use Dropout not only in the training but also in the generation phase.\\
Dropout\cite{srivastava2014dropout} is usually used in GAN layers for regularization to prevent over-fitting: units are deactivated with a given probability $p$. This is done to prevent co-adaptions betweens units. These co-adaptions prevent generalization, so unseen data performs worse.\\
Dropout in one unit $i$ is defined as:
\begin{eqnarray*}
Training: O_i &=& X_ia(\sum_{k=1}^{d_i}w_kx_k+b_i) \\
Generation: O_i &=& qa(\sum_{k=1}^{d_i}w_kx_k+b_i) 
\end{eqnarray*}
With $P(X_i = 0) = p$ and $q = 1-p$.\\
In the generation phase, the activation function $a$ is scaled by $q$ to match the expected output from the training phase.\\
\ \\
Though, most implementations use Inverted Dropout, which is defined as:
\begin{eqnarray*}
Training: O_i &=& \frac{1}{q}X_ia(\sum_{k=1}^{d_i}w_kx_k+b_i) \\
Generation: O_i &=& a(\sum_{k=1}^{d_i}w_kx_k+b_i)
\end{eqnarray*}
This slight change (scaling in the training phase instead of in the generation phase) gives the improvement, that in the generation phase no scaling or other alteration is required.\\
\ \\
In our experiments, we use Inverted Dropout for better comparability. To induce noise in the generation process, we use the same formula for testing as for generation.
\begin{eqnarray*}
Generation: O_i = \frac{1}{q^\prime}X_i^\prime a(\sum_{k=1}^{d_i}w_kx_k+b_i)
\end{eqnarray*}
But we use independent probability variables, so that scaling and dropout can be controlled separately:
$P(X_i^\prime = 0) = p_{dropout}$ and $q^\prime = 1-p_{scale}$. \\
\section{Experiment Design}
For our experiments, several models were trained on the MNIST dataset\cite{lecun1998mnist} using different probabilities $p$ for dropping out units in the training phase: 0 (which is the equivalent of using no Dropout), 0.2, 0.4, 0.6 and 0.8.\\
The GAN architecture is derived from DCGAN\cite{radford2015unsupervised}:\\
The Discriminator hidden layers consist of a 2D-Convolution, Batch Normalization and LeakyReLu. The output layer consists of a 2D-Convolution and a sigmoid activation function.\\
The Generator hidden layer consist of 2D Transposed Convolution, Batch Normalization, ReLu. We added Dropout at the end of the Sequence. The output layer consists of a 2D Transposed Convolution and hyperbolic tangent as activation function.\\
\ \\
The experiments aim to find the best Dropout configuration to both achieving the broadest variety of generated images but also the visually most appealing images.\\
To measure the variety, $N=500$ noise vectors $z$ were drawn. With these noise vectors we generate images with the unaltered generator $g(x)$ that uses no Dropout in the generation phase and an altered generator $g^\prime(x)$ that uses Dropout while generating. Between these two outputs, the euclidean distance $d$ is calculated. To minimize statistical errors, these calculations are repeated $R=100$ times. Finally the standard deviation of all distances is calculated and used as metric for variety.\\
\begin{equation*}
std(\sum_{j=1}^{R}\sum_{i=1}^{N}(d(g(z_i),g^\prime(z_i))))
\end{equation*}
We calculated the standard deviation for different settings:
\begin{itemize}
	\item Dropout applied to all hidden layers.
	\item Dropout applied to only the first hidden layer (first layer after input).
\end{itemize}
Usually Dropout is applied to all hidden layers, but in terms of generating a variety of outputs it might be interesting to apply Dropout only on the early hidden layers. This way certain features or concepts will not be used in the generation process because of the dropped out units. These missing concepts might create errors in the generated output. But if Dropout is only applied on the first layers, other layers may be able to fix these errors, which might result in an overall more consistent result.
\begin{itemize}
	\item No Scaling ($p_{scale} = 0$).
	\item Scaling matches Dropout probability\\ ($p_{scale} = p_{dropout}$).
\end{itemize}
Usually, if Dropout is applied, a unit's output is scaled by $\frac{1}{1-p}$ to match the expected output and prevent over-saturation. But in this case, the goal is to generate a variety of outputs which in the best case are a creative addition to a human-in-the-loop system. So removing the scaling but still dropping out units might give a more interesting or creative result.
\section{Results}
\label{results}
\begin{figure}[h]
	\centering
	\includegraphics[width=0.55\linewidth]{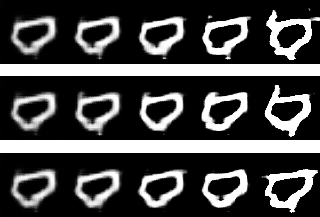}
	\caption{Model trained with a Dropout rate of 0.8. Images generated with Dropout rates ranging from 0 to 0.8. Generation was repeated three times to show variety in output. Especially with higher Dropout rates, generated images differ a lot.}
	\label{model8}
\end{figure}

Table \ref{scaler_layerall} shows the results of using typical Dropout configuration (same as in training) in the generation phase. \textit{Training $p$} states the Dropout probability that was used to train the model. So each column represents a separate model. \textit{Generation $p$} states the Dropout probability that was used for scaling ($p_{scale}$) and as Dropout probability ($p_{dropout}$).\\
\ \\
For each model (so no matter with which Dropout probability it was trained), it is clearly visible that a higher $p$ in Generation results in a larger difference between outputs. This is also visible in Figure \ref{model8} where one noise vector was 
used to generate images with different Dropout rates starting at 0 on the left and go up to 0.8 on the right. The generation series was repeated 3 times. With a higher dropout rate, the images differ more between each generation. Also conspicuous is that with a higher dropout rate, the images tend to have sharper edges. This leads to the conclusion that details are getting lost.

\begin{figure}[h]
	\centering
	\includegraphics[width=0.5\linewidth]{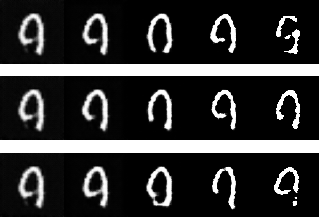}
	\caption{Model trained with no Dropout. Images generated with Dropout rates ranging from 0 to 0.8. Generation was repeated three times to show variety in output. Especially with higher Dropout rates, generated images look distorted or broken.}
	\label{nodropout}
\end{figure}

The standard deviation also increases with a larger Dropout probability in training. Except if the model was trained completely without Dropout. In this case, the standard deviation directly jumps to values similar to a Training $p$ of 0.6. Figure \ref{nodropout} shows corresponding images: If the model was trained with no Dropout but Dropout is used in generation, the resulting images look broken, especially with higher probabilities. This is most certainly due to learned co-adaptions between units. These co-adaptions are not reliable anymore if Dropout is applied, so the generation breaks.

\begin{table}[ht]
	\small
\begin{tabular}{cc|ccccc}
	\multicolumn{2}{c|}{\small{all layers}} & \multicolumn{5}{c}{\textbf{Training $p$}} \\ 
	\multicolumn{2}{c|}{\small{matching Scale}} & \textbf{0} & \textbf{0.2} & \textbf{0.4} & \textbf{0.6} & \textbf{0.8}\\
	\hline 
	\multirow{5}{*}{\rotatebox[origin=c]{90}{\textbf{Generation $p$}}} &\textbf{0}& 0& 0& 0 & 0& 0\\
	&\textbf{0.2}& 1.258 & 1.158 & 1.227&  1.342 & 1.55 \\
	&\textbf{0.4}& 2.092 & 1.716 & 1.804& 1.994 & 2.668  \\
	&\textbf{0.6}& 3.027 & 2.394 & 2.55 & 2.752 & 3.847 \\
	&\textbf{0.8}& 4.116 & 3.468 & 3.683& 3.973& 5.213 
\end{tabular} 
	\caption{Standard deviation of models tested with Dropout on all hidden layers. The scale factor matches the Dropout probability in generation ($p_{scale} = p_{dropout}$).}
	\label{scaler_layerall}
\end{table}

\subsection{Scaling}
\begin{figure}[h]
	\centering
	\includegraphics[width=0.5\linewidth]{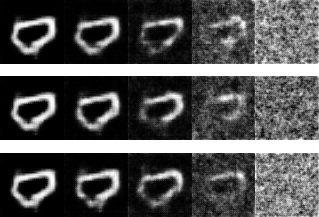}
	\caption{Model trained with a Dropout rate of 0.8. Images generated with Dropout rates ranging from 0 to 0.8 and no scaling ($p_{scale}=0$). Generation was repeated three times to show variety in output. }
	\label{scale0}
\end{figure}

In this experiment, Dropout was applied to the same models as in section \ref{results}, but no scaling was used ($p_{scale}=0$). Table \ref{scale0_layerall} shows the resulting standard deviations from all tested models and Dropout rates. The results look different than before: the variety first increase with a higher dropout rate but then shrinks again. A higher Dropout rate in training again gives the largest standard deviation.\\
Figure \ref{scale0} helps to understand why the variety decreases with high Dropout rates: With medium Dropout rates in generation, the result images look slightly different but also start to get noisy. If the Dropout rate is increased even more, the image generation completely breaks and only results in random noise. This is most certainly due to under-saturation in units: Dropout is applied, so the average signal value is decreased. Usually this value would be scaled back to match the expected output, but in this experiment no scaling is applied. So the unit's output stays at it's low level, which results in noisy images. So, values should definitely be scaled if Dropout is used.

\begin{table}[ht]
	\small
	\begin{tabular}{cc|ccccc}
		\multicolumn{2}{c|}{\small{all layers}} & \multicolumn{5}{c}{\textbf{Training $p$}} \\ 
		\multicolumn{2}{c|}{\small{no Scaling}} & \textbf{0} & \textbf{0.2} & \textbf{0.4} & \textbf{0.6} & \textbf{0.8}\\
		\hline 
		\multirow{5}{*}{\rotatebox[origin=c]{90}{\textbf{Generation $p$}}} &\textbf{0}& 0& 0& 0 & 0& 0\\
		&\textbf{0.2}& 1.204 &1.141 &1.097 &1.301 &1.733  \\
		&\textbf{0.4}& 1.544 &1.399 &1.148 &1.763 &3.233   \\
		&\textbf{0.6}& 1.713 &1.755 &1.235 &2.227 &3.441   \\
		&\textbf{0.8}& 1.126 &1.272 &0.777 &0.909 &1.517   
	\end{tabular} 
	\caption{Standard deviation of models tested with dropout on all hidden layers and no Scaling.}
	\label{scale0_layerall}
\end{table}

\subsection{Dropout and Layers}
\begin{figure}[h]
	\centering
	\includegraphics[width=0.5\linewidth]{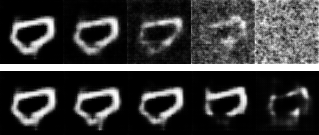}
	\caption{Model trained with a Dropout rate of 0.8. Images generated with Dropout rates ranging from 0 to 0.8 and no scaling. The first row shows a model with Dropout applied to all hidden layers in generation, the bottom row has Dropout only applied to the first hidden layer.}
	\label{repair}
\end{figure}

If Dropout is only applied in the first hidden layers, the remaining layers might be able to fix errors that emerge due to deactivated units. Therefor in this experiment, Dropout was only applied to the first hidden layer of the models described in \ref{results}. When comparing the initial experiment setting (Dropout on all hidden layers) versus Dropout only applied to the first hidden layer, no significant differences could be found in the standard deviation. Generated images also look very similar and no distinct difference could be recognized.\\
However, Figure \ref{repair} demonstrates that the mentioned repair ability of additional non-Dropout layers exists. The first row shows one row of the no-scaling experiment. A model was trained with a Dropout rate of 0.8 and tested with Dropout rates ranging from 0 to 0.8 but without scaling ($p_{scale}=0$). This results in very noisy images.\\
The second row shows the exact same setup with the difference, that the Dropout without scaling only was applied on the first hidden layer. The resulting images are less noisy and up to medium Dropout ranges, the Dropout-induced errors are visually repaired very well. On high Dropout rates, the remaining layers do not completely succeed in fixing the errors from earlier layers, but still improve the image quality.

%

\subsection{Test on Layer Mask Dataset}
\begin{figure}[h]
	\centering
	\includegraphics[width=0.95\linewidth]{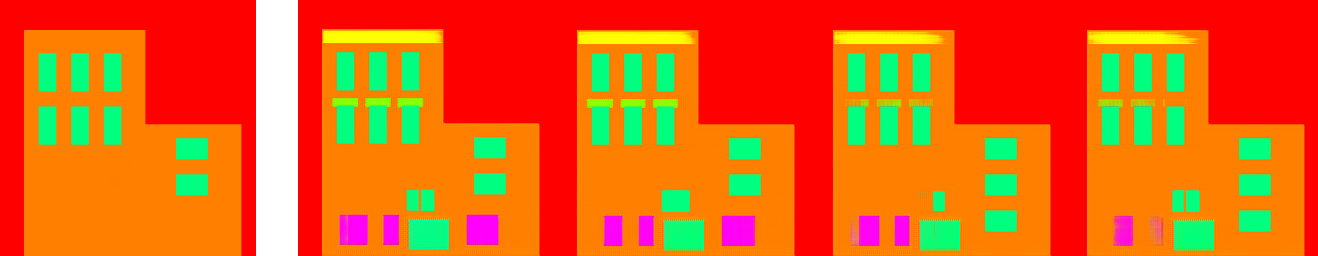}
	\caption{Model trained on facade label masks. Left images shows input, right images show variety of output if Dropout rate of 0.5 is used for generation. }
	\label{facadeex}
\end{figure}

To give a better impression how Dropout can be used to generate a variety of results, we trained a model on the label masks in the CMP Facades dataset\cite{Tylecek13}. The model was trained to add additional labels to a label mask only containing the facade and window labels. The training goal was to let the model add additional labels, so that the facade would still looks coherent. Additional labels can be other windows, shops, sills, moldings, etc. This way the model could for example help an architect to decide where to put the next facade element. The neural net architecture matches the pix2pix architecture described in \cite{isola2017image}.\\
Figure \ref{facadeex} shows the generative results using our model. It was trained with a Dropout rate of 0.5. The image on the left shows the input image and the right side shows four generated outputs using a Dropout rate of 0.5. The model adds additional windows (turquoise), shops(pink), moldings(yellow) and cornices(green). The generated images differ mainly in placing and size of shops and windows. Windows are also sometimes split into two separate ones.\\
The resulting images are coherent, give different suggestions to the user and could therefore very well be used in a human-in-the-loop system.

\section{Conclusion}
In our experiments, we showed that Dropout is a suitable method in GANs to generate a variety of outputs to one input, especially if other methods like latent space exploration cannot be used. Larger Dropout rates give a larger variety but might also result in incoherent images. The Dropout rate in training also influences the result with larger rates also resulting in a larger variety. However training with large Dropout rates is difficult and may result in mode collapse. Therefore training and generation Dropout rates should be set as high as possible with the trade-offs in mind.\\
Applying Dropout only to the first hidden layers might be beneficial to the image quality, because the non-Dropout layers can repair errors emerging due to Dropout in the early layers. How advantageous this is exactly will have to be researched in future studies.\\

{\small
\bibliographystyle{ieee}
\bibliography{egbib}

\begin{thebibliography}{10}\itemsep=-1pt

\bibitem{chen2016infogan}
X.~Chen, Y.~Duan, R.~Houthooft, J.~Schulman, I.~Sutskever, and P.~Abbeel.
\newblock Infogan: Interpretable representation learning by information
  maximizing generative adversarial nets.
\newblock In {\em Advances in neural information processing systems}, pages
  2172--2180, 2016.

\bibitem{gatys2015neural}
L.~A. Gatys, A.~S. Ecker, and M.~Bethge.
\newblock A neural algorithm of artistic style.
\newblock {\em arXiv preprint arXiv:1508.06576}, 2015.

\bibitem{hinton2012improving}
G.~E. Hinton, N.~Srivastava, A.~Krizhevsky, I.~Sutskever, and R.~R.
  Salakhutdinov.
\newblock Improving neural networks by preventing co-adaptation of feature
  detectors.
\newblock {\em arXiv preprint arXiv:1207.0580}, 2012.

\bibitem{isola2017image}
P.~Isola, J.-Y. Zhu, T.~Zhou, and A.~A. Efros.
\newblock Image-to-image translation with conditional adversarial networks.
\newblock In {\em Proceedings of the IEEE conference on computer vision and
  pattern recognition}, pages 1125--1134, 2017.

\bibitem{kingma2018glow}
D.~P. Kingma and P.~Dhariwal.
\newblock Glow: Generative flow with invertible 1x1 convolutions.
\newblock In {\em Advances in Neural Information Processing Systems}, pages
  10215--10224, 2018.

\bibitem{lecun1998mnist}
Y.~LeCun.
\newblock The mnist database of handwritten digits.
\newblock {\em http://yann. lecun. com/exdb/mnist/}.

\bibitem{mathieu2015deep}
M.~Mathieu, C.~Couprie, and Y.~LeCun.
\newblock Deep multi-scale video prediction beyond mean square error.
\newblock {\em arXiv preprint arXiv:1511.05440}, 2015.

\bibitem{mirza2014conditional}
M.~Mirza and S.~Osindero.
\newblock Conditional generative adversarial nets.
\newblock {\em arXiv preprint arXiv:1411.1784}, 2014.

\bibitem{pathak2016context}
D.~Pathak, P.~Krahenbuhl, J.~Donahue, T.~Darrell, and A.~A. Efros.
\newblock Context encoders: Feature learning by inpainting.
\newblock In {\em Proceedings of the IEEE conference on computer vision and
  pattern recognition}, pages 2536--2544, 2016.

\bibitem{radford2015unsupervised}
A.~Radford, L.~Metz, and S.~Chintala.
\newblock Unsupervised representation learning with deep convolutional
  generative adversarial networks.
\newblock {\em arXiv preprint arXiv:1511.06434}, 2015.

\bibitem{reed2016learning}
S.~E. Reed, Z.~Akata, S.~Mohan, S.~Tenka, B.~Schiele, and H.~Lee.
\newblock Learning what and where to draw.
\newblock In {\em Advances in Neural Information Processing Systems}, pages
  217--225, 2016.

\bibitem{srivastava2014dropout}
N.~Srivastava, G.~Hinton, A.~Krizhevsky, I.~Sutskever, and R.~Salakhutdinov.
\newblock Dropout: a simple way to prevent neural networks from overfitting.
\newblock {\em The journal of machine learning research}, 15(1):1929--1958,
  2014.

\bibitem{Tylecek13}
R.~Tyle{\v c}ek and R.~{\v S}{\' a}ra.
\newblock Spatial pattern templates for recognition of objects with regular
  structure.
\newblock In {\em Proc. GCPR}, Saarbrucken, Germany, 2013.

\bibitem{yeh2017semantic}
R.~A. Yeh, C.~Chen, T.~Yian~Lim, A.~G. Schwing, M.~Hasegawa-Johnson, and M.~N.
  Do.
\newblock Semantic image inpainting with deep generative models.
\newblock In {\em Proceedings of the IEEE Conference on Computer Vision and
  Pattern Recognition}, pages 5485--5493, 2017.

\bibitem{Zhang_2017_ICCV}
H.~Zhang, T.~Xu, H.~Li, S.~Zhang, X.~Wang, X.~Huang, and D.~N. Metaxas.
\newblock Stackgan: Text to photo-realistic image synthesis with stacked
  generative adversarial networks.
\newblock In {\em The IEEE International Conference on Computer Vision (ICCV)},
  Oct 2017.

\bibitem{zhu2016generative}
J.-Y. Zhu, P.~Kr{\"a}henb{\"u}hl, E.~Shechtman, and A.~A. Efros.
\newblock Generative visual manipulation on the natural image manifold.
\newblock In {\em European Conference on Computer Vision}, pages 597--613.
  Springer, 2016.

\bibitem{zhu2017unpaired}
J.-Y. Zhu, T.~Park, P.~Isola, and A.~A. Efros.
\newblock Unpaired image-to-image translation using cycle-consistent
  adversarial networks.
\newblock In {\em Proceedings of the IEEE international conference on computer
  vision}, pages 2223--2232, 2017.

\end{thebibliography}
}

\end{document}